\documentclass[twoside]{article}

\usepackage[top=2cm,bottom=3cm, inner=3cm,outer=5.5cm]{geometry} 
\usepackage{nameref}
\usepackage[comma]{natbib}
\usepackage{url}
\usepackage{amssymb,amsmath}
\usepackage{bbm} 
\usepackage{tcolorbox} 
\usepackage{hyperref}

\newtcolorbox{textbox}[2][]{colback=blue!5!white, colframe=blue!75!black, fonttitle=\bfseries, title=#2, #1}

\usepackage{marginnote}
\usepackage[strict]{changepage} 
\newcommand{\oddentry}[2]{\marginpar{\setlength{\leftskip}{0.5cm}\small\textbf{#1:} #2}}
\newcommand{\evenentry}[2]{\marginpar{\setlength{\rightskip}{0.5cm}\small\textbf{#1:} #2}}
\newcommand{\entry}[2]{\checkoddpage\ifoddpage \oddentry{#1}{#2}\else \evenentry{#1}{#2}\fi}

\newcommand{\zerob} {{\bf 0}}

\newcommand{\thetab} {{\boldsymbol{\theta}}}
\newcommand{\alphab} {{\boldsymbol{\alpha}}}

\newcommand{\nub} {{\boldsymbol{\nub}}}

\newcommand{\etab} {{\boldsymbol{\eta}}}
\newcommand{\Thetab} {{\boldsymbol{\Theta}}}

\newcommand{\varthetab} {{\boldsymbol{\vartheta}}}

\newcommand{\Xset} {\mathcal{X}}
\newcommand{\intd} {\textrm{d}}
\newcommand{\phib} {\boldsymbol{\phi}}

\newcommand{\Wmat} {\textbf{W}}

\newcommand{\Cmat} {\mathbf{C}}

\newcommand{\Imat} {\textbf{I}}
\newcommand{\Umat} {\textbf{U}}

\newcommand{\Vmat} {\textbf{V}}
\newcommand{\bvec} {\textbf{b}}

\newcommand{\hvec} {\textbf{h}}

\newcommand{\xvec} {\textbf{x}}

\newcommand{\svec} {\textbf{s}}
\newcommand{\uvec} {\textbf{u}}

\newcommand{\fvec} {\textbf{f}}
\newcommand{\rvec} {\textbf{r}}

\newcommand{\betab} {\boldsymbol {\beta}}

\renewcommand{\zerob}{\mathbf{0}}

\newcommand{\Yvec}{\mathbf{Y}}

\newcommand{\Zvec}{\mathbf{Z}}
\newcommand{\epsilonb}{\boldsymbol{\epsilon}}

\newcommand{\cov}{\mathrm{cov}}

\newcommand{\Unif}{\mathrm{Unif}}
\newcommand{\Bern}{\mathrm{Bern}}
\newcommand{\GP}{\mathrm{GP}}

\DeclareMathOperator*{\argmax}{arg\,max}

\newcommand{\res}{\textrm{res}}

\let\originalleft\left
\let\originalright\right
\renewcommand{\left}{\mathopen{}\mathclose\bgroup\originalleft}
\renewcommand{\right}{\aftergroup\egroup\originalright}

\begin{document}

\sloppy

\title{Statistical Deep Learning for Spatial and Spatio-Temporal Data}
\author{Christopher K. Wikle$^1$, and Andrew Zammit-Mangion$^2$}
\date{\small $^1$ Department of Statistics, University of Missouri, MO \\ $^2$ School of Mathematics and Applied Statistics, University of Wollongong, Australia}
\maketitle

\begin{abstract}
  Deep neural network models have become ubiquitous in recent years, and have been applied to nearly all areas of science, engineering, and industry.  These models are particularly useful for data that have strong dependencies in space (e.g., images) and time (e.g., sequences). 
  Indeed, deep models have also been extensively used by the statistical community to model spatial and spatio-temporal data through, for example, the use of multi-level Bayesian hierarchical models and deep Gaussian processes. In this review, we first present an overview of traditional statistical and machine learning perspectives for modeling spatial and spatio-temporal data, and then focus on a variety of hybrid models that have recently been developed for latent process, data, and parameter specifications. These hybrid models integrate statistical modeling ideas with deep neural network models in order to take advantage of the strengths of each modeling paradigm. We conclude by giving an overview of computational technologies that have proven useful for these hybrid models, and with a brief discussion on future research directions.
 \end{abstract}


\section{Introduction}\label{sec:intro}
Deep learning has revolutionized prediction and classification for many types of dependent data.  These dependencies are often temporal and spatial in nature, and  state-of-the-art  modifications of recurrent and convolutional neural networks have been especially adept at accounting for this structure, at least when there are copious amounts of training data.  Deep learning models have therefore been extensively used, with great success, for natural language processing and image classification (see the sidebar titled The Deep Learning Revolution). For the most part, these models have been developed independently of models in statistics to account for such dependencies (e.g., Gaussian process (GP) models, and dynamic models; see \cite{Wikle_2019} for a recent overview of spatio-temporal statistical models).  At first glance, the neural network models used in machine learning and the statistical approaches for spatio-temporal data may seem quite distinct. However, in complex data applications, both approaches tend to rely on multi-level (hierarchical) representations, with the primary differences being in the types of applications for which they are used, and in the fact that the statistical approaches for multi-level models tend to accommodate uncertainty quantification more naturally within a probabilistic framework.  

There has been an increasing number of works in recent years, especially in the statistics literature, that take a hybrid approach to modeling complex spatial or spatio-temporal data.  These hybrid models are predominantly classical hierarchical statistical models that borrow some of the effective ideas from deep neural models in order to facilitate modeling the data, process, and/or parameter components that form the hierarchy. These hybrid models have led to many notable advances in statistical spatial/spatio-temporal modeling and inference. For example:

\begin{itemize}  \setlength\itemsep{1em}
\item \citet{Siden_2020} use deep learning to construct a flexible family of Gaussian Markov random fields, subsequently used to model and predict land surface temperatures,\footnote{\url{https://github.com/finnlindgren/heatoncomparison}} while \citet{Chen_2021} integrate deep learning with the basis function approach to modeling spatial processes to predict PM$_{2.5}$ concentrations across the United States. Both methods are shown to be superior to more classical approaches to spatial prediction.
  \item  \citet{Lenzi_2021} use deep learning to estimate parameters with statistical spatial models of extremes, whose likelihood is intractable or difficult to evaluate. They apply their methods for efficiently estimating parameters in a Brown--Resnick process fitted to surface temperature data.\footnote{\url{https://ldas.gsfc.nasa.gov/nldas/v2/models}} \citet{Zammit_2020} show how deep learning methods can be used to estimate spatially and temporally varying dynamics of a statistical spatio-temporal model, and provide uncertainty quantification on these dynamics, at a fraction of the computational cost that would typically be required. They apply their methods to efficiently forecast sea-surface temperature, using data available from the Copernicus Marine Environment Monitoring Service (CMEMS),\footnote{\url{https://marine.copernicus.eu/}} and for nowcasting rainfall, using radar reflectivity data available from \citet{STRbook}.
\item \cite{mcdermott2017ensemble} show how  spatio-temporal echo state networks (ESNs)  can be used to produce long-lead forecasts and uncertainty quantification of tropical Pacific Ocean sea surface temperature (STT) anomalies that suggest developing El Ni\~no or La Ni\~na  conditions.  The SST data are available from \citet{STRbook}. \cite{McDermott_2019b} also show how deep versions of ESNs can be used for long-lead spatio-temporal prediction of soil moisture over the ``corn belt’’ region of the U.S. given SST data in previous months. They use SST data from the extended reconstruction SST (ERSST) data set\footnote{\url{http://iridl.ldeo.columbia.edu/SOURCES/.NOAA/.NCDC/.ERSST/}} and soil moisture data from the Climate Prediction Center's high resolution monthly global soil moisture data set.\footnote{\url{https://iridl.ldeo.columbia.edu/SOURCES/.NOAA/.NCEP/.CPC/.GMSM/.w/}} 
ESNs can also be used effectively for short-term forecasts of complex processes. \cite{huang2021forecasting} use a spatio-temporal ESN to forecast short-term wind speeds for power production over Saudi Arabia with high-resolution wind fields.\footnote{\url{https://github.com/hhuang90/KSA-wind-forecast}}

\end{itemize}

The above are just some of the highlights that will be expanded on in this review in Sections 3 and 4, following an overview of machine learning and statistical approaches for spatial and spatio-temporal data in Section 2. The review discusses technologies that enable deep learning for spatial and spatio-temporal data in Section 5, and concludes with a brief discussion of future research directions in Section 6.

\section{Conventional statistical and  machine learning approaches for spatial and spatio-temporal data}\label{sec:tradAI}

We begin the review with a brief overview of some of the key statistical and machine learning approaches for analyzing spatial and spatio-temporal data. 

\begin{textbox}{The deep learning revolution}
  A deep learning model is a statistical or machine learning model that encompasses multiple connected components to classify or predict with complex data sets. 
  There have been remarkable success stories reported in popular media outlets about how deep learning algorithms have mastered complex games such as Go, Chess, or Shogi.  More generally, these methods have proven exceptionally effective at classifying image and sequence (e.g., natural language) data because they are particularly suited for learning patterns in complex data.  Indeed, deep learning models are today at the heart of many devices and applications that are in everyday use, such as smart phone applications, web searching, and advertising, to name just a few.  Although these models have proven useful, they are not without their flaws.  For example, they typically require a great deal of training data and computational resources to train, and they have been known to fail spectacularly on occasion.  Perhaps more troubling is that most deep learning algorithms are ``black boxes'' and it is difficult to know why they are producing the prediction or classification that they do, and there is usually no measure of uncertainty associated with those outputs.  The first issue is problematic as it can lead to models that are learning unknown, and possibly unpredictable, biases. The second issue is problematic when one is attempting to use the output from these models to make decisions -- it is challenging to make decisions based on model output without knowing how reliable the output is.  For these reasons, recent years have seen an increased and sustained effort by machine learners and statisticians to understand these models, produce uncertainties for model outputs, and combine these deep neural methods with more traditional multi-layer statistical models.
\end{textbox}

\subsection{Statistical approaches for spatial data}\label{sec:tradspace}

Statistical methods for spatial data are well summarized in a variety of monographs; see, for example, \citet{cressie1993statistics} and \citet{banerjee2014hierarchical}.
These methods have historically been grouped by the support of the data they are used to model.  {\it Geostatistical} spatial methods are generally used with responses that are available at a set of point-referenced locations in a domain of interest $G$, where we term $G$ the ``geographic domain''. {\it Areal} or {\it lattice} spatial methods are often used with responses that are defined over a finite number of (generally non-overlapping) partitions of $G$. \emph{Spatial point process} methods are used when the data are a finite set of random locations in $G$ (indicative of presence/absence). 
Other types of spatial data include random sets and trajectories. 

Assume we can write our random spatial process as
\begin{equation}
Y(\svec) = f(\svec; \betab) + \eta(\svec), \quad \svec \in G,
\label{eq:Yprocess}
\end{equation}
where $f(\cdot\,; \betab)$ is the conditional mean of $Y(\cdot)$ that contains covariate relationships and associated fixed effects $\betab$ (e.g., $f(\cdot\,;\betab) = \xvec'(\cdot)\betab $, where $\xvec(\cdot)$ is a set of known covariates or ``features''), and $\eta(\cdot)$ is the spatially-dependent random process. In the context of geostatistical models, $\eta(\cdot)$  is often modeled as  a Gaussian process (GP).  A GP is a dependent process where all of its finite-dimensional distributions are Gaussian and defined through a mean function $\mu(\cdot)$ and a covariance function $C(\svec,\tilde{\svec}) = \cov(Y(\svec),Y(\tilde{\svec}))$ where $\svec,\tilde{\svec} \in G$. A GP $\eta(\cdot)$ with mean $\mu(\cdot)$ and covariance function $C(\cdot,\cdot)$ is denoted as $\eta(\cdot) \sim \GP(\mu(\cdot),C(\cdot,\cdot))$.  In geostatistical applications, it is typically the case that $\mu(\cdot) = 0$ since the conditional mean is accounted for by $f(\cdot;\betab)$ (note, this implies that $Y(\cdot) \sim \GP(f(\cdot;\betab),C(\cdot,\cdot))$).

Perhaps the biggest challenge in implementing GP-based spatial prediction has to do with the covariance function, $C(\cdot,\cdot)$. In practice, this is not known, and one must typically assume stationarity (intrinsic or second-order) and often isotropy (directional invariance) to proceed. In addition, the functional form of the stationary covariance matrix is often specified (e.g., Gaussian, Mat\'ern, power) up to some unknown parameters, $\thetab_y$ say.  Even in these cases, the form of the optimal predictor requires that one evaluates the inverse of the covariance matrix associated with all observation locations, and this can be problematic in high-data-volume problems.    \entry{Covariance stationarity}{a covariance function $C(\svec,\cdot), \svec \in G$, is stationary if it only depends on the displacement from $\svec$, that is, if one can write $C(\svec, \uvec) = C^o(\hvec)$ for any $\svec, \uvec \in G$, where $\hvec \equiv \svec - \uvec$ \vspace{0.1in}}
    \entry{Covariance isotropy}{a covariance function $C(\svec,\cdot), \svec \in G$, is isotropic if it only depends on the distance from $\svec$, that is, if one can write $C(\svec, \uvec) = C^d(\svec, d)$ for any $\svec, \uvec \in G$, where $d \equiv \|\svec - \uvec\|$\vspace{0.1in}}
Much of the research in spatial statistics over the last decade has been concerned with developing methods for such high-data-volume prediction problems; these approaches have tended to fall into two categories --  neighbor-based methods and basis function approaches \citep[see][for a recent overview]{heaton2019case}.

When dealing with areal or lattice data, the random component in Equation \ref{eq:Yprocess} is typically modeled as a Gaussian Markov random field (MRF).  In this case, one often is interested in making inference on the conditional mean parameters, or in smoothing noisy areal observations (e.g., as in disease mapping). Gaussian MRFs lead to a highly structured and yet parsimonious sparse precision matrix, which can facilitate computation in marginal formulations or conditional (Bayesian) implementations.

   \entry{Gaussian Markov random field}{a collection of variables that are jointly Gaussian, and where one or more variables is conditionally independent of others when conditioned on variables in a neighboring set \vspace{0.1in}}

   \entry{Hierarchical Model}{a model constructed via a series of conditional distributions and marginal distributions (e.g., $[A,B,D] = [D\mid A,B][A \mid B][B]$) \vspace{0.1in}}

   \entry{Bayesian Hierarchical Model (BHM)}{a hierarchical model where Bayes' rule is used to make inference on all unknown quantities given data.  For example, if $D$ represents data, $[A,B | D] \propto [D | A,B][A|B][B]$ \vspace{0.1in}}

   Regardless of whether one treats $Y(\cdot)$ from a GP or an MRF perspective, it is best to treat it as a latent process, which is only partially observed via a finite set of $m$, say, spatial observations, $\Zvec \equiv (Z_i: i = 1,\dots,m)'$, where each $Z_i$ is an observation made at $\rvec_i \in G$, or averaged over $\rvec_i \subset G, i = 1,\dots,m$. Then, a model can be specified for the observations conditional on this latent process, $[\Zvec \mid Y(\cdot),\thetab_z]$ say, where the brackets $[\;\cdot \;]$ denote a probability distribution, and $\thetab_z$ here denotes parameters that parameterize the conditional distribution of the data. As in generalized linear mixed models, this model easily accommodates non-Gaussian observations as well as measurement error. Although the parameters associated with the data model and latent process model, $\{\thetab_z,\thetab_y\}$, can be estimated through likelihood methods, it is often the case that one specifies prior distributions for these parameters and considers a Bayesian implementation \citep[e.g., see][]{cressie2011statistics, banerjee2014hierarchical}.  This leads to a multi-level Bayesian hierarchical model (BHM):
\begin{eqnarray*}
\mbox{Data Model:}  & \; & [\Zvec \mid Y(\cdot), \thetab_z], \\
\mbox{Process Model:} &\; & [Y(\cdot) \mid f(\cdot\,; \betab),  \thetab_y], \\
\mbox{Parameter Models:} &\; & [\thetab_z, \thetab_y, \betab].
\end{eqnarray*}
This BHM formulation is important as each model component is easily expanded to account for more complexity such as multiple data sources, multivariate processes, and parameters that are themselves processes (e.g., spatially-varying fixed effects).  Such multi-level models are very much ``deep'' models as we see in the remainder of the review.

 \subsection{Statistical approaches for spatio-temporal data}\label{sec:tradST}

Statistical methods for spatio-temporal data are extensively described in the monographs of \citet{cressie2011statistics} and \citet{Wikle_2019}. Such methods can also be classified according to the spatial support of the data they are used to analyze (i.e., geostatistical, lattice, point process, etc.) and primarily differ from their spatial counterparts through the inclusion of a time index, which is assumed to come from a discrete set or a continuum.  We represent the spatio-temporal process as $\{Y(\svec;t): \svec \in G, t \in \mathcal{T}\}$ where $t$ indexes the temporal domain $\mathcal{T} \subset \mathbb{R}^1$. As with the purely spatial process, we can consider either a GP or an MRF representation for the process $Y(\svec;t) = f(\svec, t; \betab) + \eta(\svec;t)$. Now the dependence is given by a spatio-temporal covariance function, say, $C(\svec,t;\tilde{\svec},\tilde{t}) \equiv \cov(Y(\svec;t),Y(\tilde{\svec},\tilde{t}))$.  
In the GP case, the same covariance specification challenges occur as with the purely spatial case in terms of stationarity and high dimensionality (both of which are more challenging in the spatio-temporal case). The spatio-temporal case is further complicated by the difficulty in specifying realistic joint covariances.  For this reason, and to facilitate computation, spatio-temporal covariance functions are often assumed to be separable, that is, $C(\svec,t;\tilde{\svec},\tilde{t}) = C_s(\svec,\tilde{\svec}) \cdot C_t(t,\tilde{t})$.  The extra complexity of spatio-temporal data make the multi-level BHM representation even more appealing in practice.

The GP approach to spatio-temporal modeling is often called a ``descriptive'' approach since it does not explicitly account for the mechanisms that generated the data. The alternative paradigm is a ``dynamical'' approach, which specifies models describing the evolution of the spatial process with time, thereby attempting to follow the etiology of the data generating mechanism.  Dynamic process models typically make a Markovian assumption in time; for example, for a discrete-time spatio-temporal model with unit time intervals, a first-order Markov assumption states that the process at time $t+1$ is independent of the process at times $t-1, t-2, \ldots$ given the process at time $t$.  Perhaps the most used dynamic spatio-temporal model (DSTM) is based on the \emph{integro-difference equation} \citep[e.g.,][Chapter 5]{Wikle_2019}:
\begin{equation}\label{eq:DSTM_IDE}
Y_{t+1}(\svec) = \int_G k(\svec, \rvec; \thetab_{k,t})Y_t(\rvec) \intd\rvec + \eta_t(\svec), \quad \svec \in G,
\end{equation}
where $t = 1,2,\dots$, denotes discrete time (note, we typically use subscripts to index time with discrete-time processes), $G$ is the spatial domain over which the process is evolving, $k(\cdot,\cdot\,; \thetab_{k,t})$ is the mixing or transition kernel, $\{\thetab_{k,t}\}$ are temporally-evolving parameters appearing in the mixing kernel, and $\eta_t(\cdot)$ is an additive, Gaussian, spatial disturbance (typically with mean zero and temporally uncorrelated).  Such models can easily be parameterized to incorporate mechanistic spatio-temporal behavior (e.g., diffusion and advection), and when discretized in space become vector autoregressive processes.  More complex spatio-temporal dynamics can be accommodated by considering nonlinear dynamic models \cite[such as quadratic nonlinear dynamic models; see][]{wikle2010general}.  As with the GP and MRF approaches, the DSTM can be used with non-Gaussian observations, in which case the dynamical process is treated as latent. It is natural to use the multi-level BHM framework to specify these models.  \citet{wikle2019comparison} presents a prototypical ``deep'' DSTM that includes layers for the data model, the conditional mean, the process model (composed of a dynamic and non-dynamic component), the dynamic process model, the non-dynamic process model, the prior distributions (that act as regularizers), and the hyperprior distributions (for a total of seven levels).  This depth is not unusual in complex spatio-temporal data applications.

\subsection{Machine learning/ai methods for spatial and spatio-temporal data}\label{sec:tradAImethods}

Many major success stories in deep learning involve spatially dependent data (e.g., image classification) and sequence data (e.g., natural language processing and time series forecasting). While classical multilayer perceptrons have been used for spatial prediction \citep[e.g.,][]{Wang_2019}, recent successes owe much to the use of structured deep networks, such as convolutional neural networks (CNNs) and recurrent neural networks (RNNs),  which have architectures particularly suited for the problem at hand. \citet{fan2021selective} present a comprehensive tutorial overview of these and other classical deep neural models from a statistician's perspective.  Not surprisingly, soon after the development of these effective CNN and RNN models, they were used to model spatio-temporal data with slight modifications \citep[e.g.,][]{wang2016cnn}. More elaborate variants of the vanilla neural networks, such as autoencoders, generative adversarial networks, tensor networks, sequence-to-sequence networks, and graph neural networks, to name a few, have also been used to model spatio-temporal data \citep[e.g.,][]{oh2015action, xingjian2015convolutional, yu2017spatio, bai2019stg2seq, guo2019deep, song2020spatial}. \entry{Convolutional Neural Network (CNN)}{a deep neural network that takes image inputs and that learns shared filters that are convolved across the image to feed into the next layer; primarily used in image classification and computer vision for learning important multi-resolution features at each hidden layer\vspace{0.1in}}
\entry{Recurrent Neural Network (RNN)}{a deep neural network that takes sequential or time series data as input, and that learns patterns in the data while accounting for “memory”; primarily used for natural language processing (e.g., translation, speech recognition) and time-series forecasting\vspace{0.1in}}

As powerful as these artificial intelligence (AI) spatio-temporal methods are, they can present limitations in the context of statistical modeling for spatial and spatio-temporal data.  For example, there is often substantial uncertainty in these types of data, including data gaps (as with satellite data), spatial or temporal supports that are at odds with the desired prediction supports, and substantial sampling and measurement uncertainty. 
Traditional deep model implementations do not directly provide model-based estimates of prediction and/or classification error that can account for these sources of error, nor can they easily incorporate or enforce known mechanistic relationships that are often present in spatio-temporal data. Further, given that these methods are essentially complex black-boxes, they are unable to perform inference or even provide guidance as to which inputs are important in explaining/predicting the response. \citet{reichstein2019deep} gives an insightful in-depth discussion of many of these issues.

 \entry{Autoencoder}{a deep learning architecture that consists of an encoder segment that finds a low dimensional representation of the data, and a decoder segment that reconstructs the data from its low dimensional representation.}

The AI community is, however, rising to these challenges.  Increased interest in uncertainty quantification and explainability is the impetus behind the so-called ``explainable AI (XAI)''  \citep[e.g.,][]{gunning2019xai} and ``interpretable AI'' \citep{rudin2022interpretable} movements.  In the context of uncertainty quantification (UQ), some success has been reported through four key approaches: variational Bayesian inference, Monte Carlo dropout, mixture density networks, and ensemble techniques \citep[see ][for an extensive overview and references]{abdar2021review}. In the context of explainability, the main aim is to ensure that models are transparent so that deep learning models do not include unanticipated biases.  Approaches to increase explainability include the use of interpretable local surrogates (such as the Local Interpretable Model-Agnostic Explanations (LIME) approach), occlusion analysis (e.g., Shapley values, SHapley Additive exPlanations (SHAP), Kernel SHAP, meaningful perturbation), integrated gradients (e.g., SmoothGrad), and layerwise relevance propagation. There are also self-explainable models (e.g., that utilize attention mechanisms) and specialized models that allow interpretability (e.g., graph neural networks).    \entry{Local Interpretable Model-Agnostic Explanations (LIME)}{an approach to model explainability that uses local (usually linear) surrogate models to explain the features that most influence individual predictions\vspace{0.1in}}
   \entry{Shapley Values}{developed in game theory to assign credit to players, depending on their contribution to the total payout from a game; now also used in AI to quantify which features are most important for a predicted response }
 \cite{molnar2022interpretable} and \citet{samek2021explaining} provide definitions and comprehensive overviews of these methods.  Many of the favored explainability approaches are model agnostic, meaning that they can be applied to essentially any predictive/classification model regardless of architecture (e.g., interpretable local surrogates and Shapley values) and could likely be used more broadly in statistical modeling.

The ubiquitous CNN-based methods used for image-type data are often not appropriate for continuous spatial processes, which are of most concern in ``geostatistical'' applications. These applications typically require optimal interpolation, and thus require the ability to predict at any location in space and provide a model-based assessment of uncertainty.  The long-standing optimal linear prediction approach to this problem is kriging (and its variants), which is based on Gaussian process theory \citep[e.g., see][]{cressie1993statistics}.  As discussed below in Section \ref{sec:DeepStat}, many of the hybrid statistical deep-learning approaches that afford UQ and continuous prediction are based on deep GPs.  A recent alternative that lies somewhere in-between the hybrid GP approaches and the CNN image approach is given by \citet{Kirkwood_2020}.  Here, they consider nonlinear functions of gridded covariates and point-level information (location/elevation data) in a CNN, and Monte Carlo dropout to do spatial prediction at any location in the domain of interest. The modeled output is the mean and variance of a normal distribution (as in the mixture density network approach to output uncertainty), but they also use Monte Carlo dropout as a Bayesian approximation \citep{gal2016dropout} to account for model uncertainty.  Alternatively, \citet{amato2020novel} decompose spatio-temporal processes in terms of temporal basis functions and stochastic spatial coefficients. This is a common approach in spatial and spatio-temporal statistics \citep[see][]{Wikle_2019}, but the difference here is that the model for the spatial coefficients is specified in terms of a set of regressions based on spatial covariates, which are then trained via a deep feedforward neural network.

In the context of spatio-temporal modeling, as mentioned in Section \ref{sec:tradspace}, one can take a GP approach as well, but it is often more realistic to consider a dynamic model.  The aforementioned hybrid CNN/RNN models can do this, albeit typically with no UQ, little interpretability, and no restrictions to enforce known mechanistic (physical/biological) relationships.  Regarding the latter, there has recently been a flurry of interest in enforcing such restrictions in deep models. 
Generally, one can encourage known dynamical constraints by adding an appropriate penalization term to the objective function and then proceeding as usual using stochastic gradient descent \citep[e.g.,][]{raissi2020hidden, wu2020enforcing, momenifar2022physics}.  Such approaches amount to a “soft constraint” and the solutions are not guaranteed to be physically consistent. This can present a problem in applications that require certain balance relations (e.g., continuity, conservation of mass, etc.). Recently, there has been progress in enforcing “hard” mechanistic constraints in deep models. For example, \citet{mohan2020embedding} consider a two-stage model in which the first stage uses an unconstrained CNN-type model to obtain a potential surface. This surface is then fed into an untrained physical model that performs appropriate transformations of the potential surface to obtain quantities of interest (e.g., velocities) in a manner that is physically consistent.  Other approaches, which take multiple model types and connect them such that physical constraints are enforced in one “physics” model component, appear in \citet{reichstein2019deep}, \citet{chattopadhyay2021towards}, and \citet{huang2021st}. This work is quite promising, but has yet to be integrated into a framework that provides model-based UQ and explainability.

\section{Hybrid deep learning spatial and spatio-temporal hierarchical models}\label{sec:DeepStat}

The models in Section~\ref{sec:tradAI} are well-suited for some applications involving spatial/spatio-temporal data where prediction is the end goal and where model interpretability and uncertainty quantification have a lower priority. In applications were the latter two properties of the model and the predictions are important, the fusion of deep learning methods with classical statistical models offers an attractive way forward. Such approaches generally involve first constructing classical spatial and spatio-temporal probability models as outlined in Sections~\ref{sec:tradspace} and~\ref{sec:tradST}, and then integrating deep learning for characterizing some, or all, of the conditional distributions. The resulting models inherit the best of both worlds: the interpretability and uncertainty quantification properties commonly associated with statistical models, and the complexity of input/output relationships commonly associated with deep learning models.  In this section we look at several ways in which deep learning has been used to some success in this way.

\subsection{Deep learning for characterizing complex process models}

Although the archetypal spatial/spatio-temporal hierarchical model consists of three layers,  it is often the process model that is the hardest to characterize: the process model generally embodies physical, chemical, ecological or biological principles that are often simplified for analytical or computational tractability. On the other hand, measurement processes are often well understood, while parameter models generally embody expert judgment on low-dimensional quantities that are relatively straightforward to construct once correctly elicited. It is therefore not a surprise that most effort in this area has focussed on integrating deep learning in the process model.

Generally, deep learning models are treated as ``black-box'' models that map the inputs (in this context, the spatial or spatio-temporal coordinates) to the outputs (observed quantities). Such models \citep[e.g.,][]{Calandra_2016} are rarely seen in the statistics literature, largely because the low-dimensional setting typically encountered in spatial applications creates some challenges that effectively preclude their use; see \citet{Duvenaud_2014} and \citet{Dunlop_2018} for insights. As a result, deep learning models in spatial and spatio-temporal statistics tend to contain quite a lot of structure, as we show in this sub-section.  

\subsubsection{Warping space to model nonstationary covariances}\label{sec:space_warping}

We first consider the most direct way to incorporate deep structures in spatial statistical models: that of applying a (structured) deep learning model on the spatial coordinates themselves.

Consider, for ease of exposition, a mean zero spatial process $Y(\svec), \svec \in G$, where the ``geographic domain'' $G \subset \mathbb{R}^2$. As discussed in Section \ref{sec:tradspace}, typically $Y(\cdot)$, is assumed to be Gaussian and stationary. The property of covariance stationarity can be relaxed by adopting the ``deformation'' approach of \citet{Sampson_1992}, whereby one first formulates a ``warping function'' $\fvec: G \rightarrow D, D \subset \mathbb{R}^2$, and then assumes the relationship,
$$\cov(Y(\svec),Y(\uvec)) \equiv C_G(\svec,\uvec) = C^o_D(\|\fvec(\svec) - \fvec(\uvec)\|),$$
for $\svec,\uvec \in G$, where $C_G(\cdot,\cdot)$ is a nonstationary covariance function on $G$, and $C_D^o(\cdot)$ is a stationary covariance function on the ``warped domain'', $D$.

Various approaches have been used for modeling $\fvec(\cdot)$: \citet{Sampson_1992} represented $\fvec(\cdot)$ using smoothing splines, \cite{Smith_1996} used basis functions derived from thin-plate splines, \citet{Snoek_2014} used beta cumulative density functions, and \citet{Schmidt_2003} used a bivariate Gaussian process. In a deep learning setting, one expresses the warping function as the composition $\fvec(\cdot) = \fvec_{n-1} \circ \cdots \circ \fvec_1(\cdot)$, where $\fvec_l(\cdot), l = 1,\dots,n-1$, are in themselves simple, elementary functions. To our knowledge, the first to adopt this approach in a spatial statistics setting were \citet{Perrin_1999}. Their approach was subsequently coined the ``deep compositional spatial model'' and extended to include other elementary warping functions, technologies often used in deep learning (see Section~\ref{sec:tech}), and spatio-temporal and multivariate dependencies, by \citet{Zammit_2021}, \citet{Vu_2022b}, and \citet{Vu_2021}. These approaches based on function composition tend to ensure that the function $\fvec(\cdot)$ is injective by constraining each of the elementary functions to be itself injective. Injectivity guarantees that space does not fold on itself after warping, which is often seen as undesirable or unphysical in many spatial and spatio-temporal applications. 
A downside of this constraint is that the elementary warping functions need to be highly structured for injectivity to be ensured, and can limit the type of warpings that can be constructed. There are ``black-box'' deep learning architectures known as ``normalizing flows'' that are injective maps by design \citep[e.g.,][]{Rezende_2015}; whilst they are largely suited for density estimation, they have also started to see some use in spatial statistics (e.g., see Sections~\ref{sec:nested} and \ref{sec:pp}).

 \entry{Space folding}{occurs when there exists $\svec,\uvec \in G$ such that $\fvec(\svec) = \fvec(\uvec)$, where $\fvec(\cdot)$ is the warping function}

\begin{figure}[t!]
  \includegraphics[width=\linewidth]{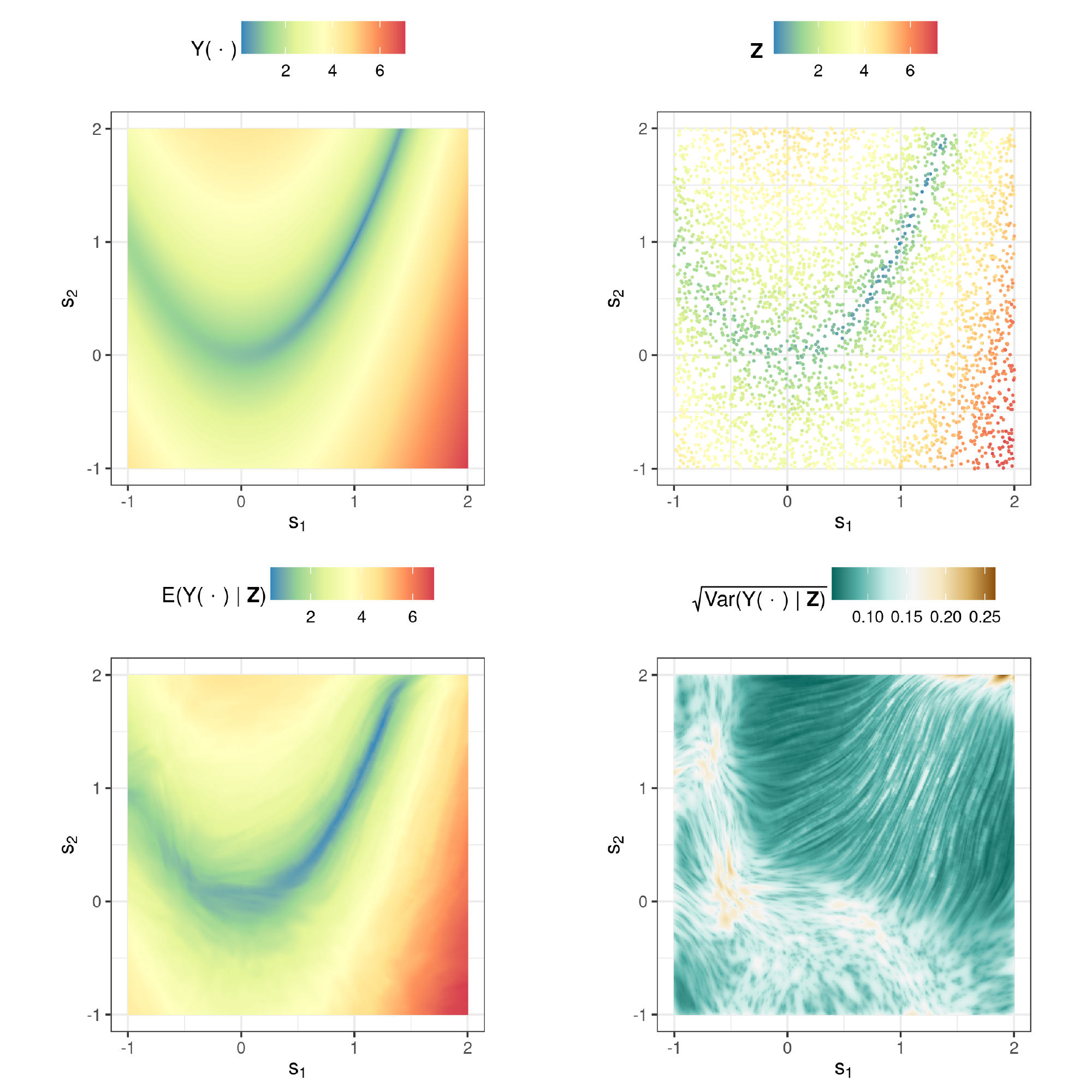}
  \caption{Illustration of spatial prediction using deep compositional spatial models. The top-left panel depicts the true underlying process, generated on $G = [-1,2] \times [-1,2]$ via Equation \ref{eq:Rosenbrock}. The top-right panel depicts the observations of the true process, that are used to train the deep spatial model. The bottom-left and bottom-right panels depict the prediction (conditional expectation) and standard error (square root of the conditional variance) on $G$ after fitting the deep model using maximum likelihood. \label{fig:Rosenbrock}}
\end{figure}

In Figure~\ref{fig:Rosenbrock} we show the utility of such models. The top-left panel depicts an underlying process, which is constructed as a variation of the Rosenbrock function, which we define on $G = [-1,2] \times [-1,2]$ as
\begin{equation}\label{eq:Rosenbrock}
 Y(\svec) = ((1 - s_1)^2 + 100(s_2 - s_1^2)^2)^\frac{1}{4}, \quad \svec \in G.
\end{equation}
The top-right panel depicts observations of this underlying process, which are point-referenced, incomplete, and noisy. The bottom-left panel shows the prediction from a deep compositional spatial model with 18 layers, where fitting was done using maximum likelihood techniques. Each layer in the model corresponds to a radial basis function \citep{Perrin_1999} that injectively warps (expands or contracts) a part of the domain; the bottom-right panel shows the associated prediction standard errors. Note how the warping captures the spatially-varying anisotropy and variability of the process, with  the valley and plateau areas of the modified Rosenbrock function readily apparent in the standard error surface.
 
 \entry{Rosenbrock function}{The classic Rosenbrock function is given by $Y(\svec) = (c_1 - s_1)^2 + c_2(s_2 - s_1^2)^2$, where $c_1, c_2 \in \mathbb{R}$}

Several of the above techniques have been considered in a Bayesian setting. One may simply equip the weights in a deep neural network with prior distributions \citep[e.g.,][Chapter 1]{Neal_1996}, and this was the approach adopted by \citet{Zammit_2021}. Alternatively, one can express the outputs at intermediate layers as Gaussian processes over the outputs of the previous layers. This construction leads to what was coined the \emph{deep Gaussian process} in the machine literature by \citet{Damianou_2013}, but which made an appearance in the spatial statistics literature a decade earlier \citep{Schmidt_2003}. The main contribution of \citet{Damianou_2013} was the use of sparse Gaussian processes \citep{Quinonero_2005} and the development of a variational inference scheme for the deep Gaussian process, that allows it to be fitted with large data sets; their architecture has been used to success in a variety of applications \citep[e.g.,][]{Salimbeni_2017}.

Deep Gaussian processes play a central role in the construction of general deep spatial models, as we discuss next.

 \entry{Deep Gaussian process}{a process constructed as $Y(\cdot) = Y_n(Y_{n-1}(\cdots (Y_1(\cdot))))$, where $Y_1(\cdot)$,$\dots$,$Y_n(\cdot)$ are Gaussian processes}

\subsubsection{Nested spatial processes}\label{sec:nested}

Models that successively warp space can be viewed as special cases of ``nested spatial processes'' where one constructs models by defining a chain of conditional probability models \citep[see][for a general framework]{Dunlop_2018}. 

\citet{Bolin_2011} consider the case where the probability models are constructed through stochastic partial differential equations via the nesting
\begin{align*}
  \mathcal{L}_nY_n(\cdot) &= Y_{n-1}(\cdot), \\
  \mathcal{L}_{n-1}Y_{n-1}(\cdot) &= Y_{n-2}(\cdot), \\
  \vdots \quad                 &~~~ \quad \vdots \\
  \mathcal{L}_2Y_2(\cdot) &= Y_1(\cdot), \\
  \mathcal{L}_1Y_1(\cdot) &= \mathcal{L}_WW(\cdot),
\end{align*}
where $\mathcal{L}_1,\dots,\mathcal{L}_n$ and $\mathcal{L}_W$ are linear operators, $W(\cdot)$ is spatial white noise, $Y_1(\cdot),\dots,Y_{n-1}(\cdot)$ are intermediate process layers, and $Y(\cdot) \equiv Y_n(\cdot)$ is the (often latent) process of interest.
A straightforward substitution yields the model $\mathcal{L}_1\cdots\mathcal{L}_nY_n(\cdot) = \mathcal{L}_WW(\cdot)$ which, under certain choices for the operators, can be easily  discretized to obtain (possibly nonstationary) Gaussian MRFs (GMRFs) with highly flexible covariance matrices. The models of \citet{Siden_2020}, coined deep GMRFs, are of similar form, but specifically designed to take advantage of CNNs that are readily available in deep learning software libraries. Their approach only considers stationary probability models and yet is seen to outperform many state-of-the-art spatial prediction methods.  Both \citet{Bolin_2011} and \citet{Siden_2020} focus on models that are nested linearly \citep[nonlinearity is briefly considered by][]{Siden_2020}, resulting in linear, Gaussian models that can subsequently be fitted using  likelihood, variational, or Markov chain Monte Carlo (MCMC) techniques.

 \entry{Variational inference}{an approximate Bayesian inference technique where the objective is to maximize a lower-bound on the marginal likelihood}

\citet{Maronas_2021} consider the related model,
\begin{align*}
  Y_n(\cdot) &= f_\varthetab (Y_1(\cdot)), \\
  Y_1(\cdot)  &\sim \GP(\mu(\cdot), C(\cdot,\cdot)),
\end{align*}
where $f_\varthetab(\cdot)$ is constructed using compositions of Sinh-Arcsinh transforms as in \citet{Rios_2019} (see also Section~\ref{sec:data_model}), but where the parameters of the transformations are themselves outputs of neural networks:
\begin{align*}
f_{\varthetab(\cdot)}(Y_1(\cdot)) &= \tilde{f}_{\varthetab_{n-1}(\cdot)} \circ \cdots \circ \tilde{f}_{\varthetab_1(\cdot)}(Y_1(\cdot)), \\
\varthetab_l(\cdot) &= \mathrm{NN}(\cdot\,, \Wmat_l),\quad l = 1,\dots,n-1,
\end{align*}
where $\mathrm{NN}(\cdot\,,\Wmat)$ denotes an arbitrary neural network with weights $\Wmat$ that takes as inputs the spatial coordinates as well as possibly other covariates of interest at the corresponding locations; $\tilde{f}_{\varthetab_l(\cdot)}(\cdot)$ is the Sinh-Arcsinh transform with parameters $\varthetab_l(\cdot)$; $\Wmat_l, l = 1,\dots,n-1,$ denote neural network weights; and where $\varthetab(\cdot) \equiv (\varthetab_1(\cdot)',\dots, \varthetab_{n-1}(\cdot)')'$ are now input (i.e., spatially) dependent transformation parameters.  \citet{Maronas_2021} develop computationally efficient techniques for estimation and prediction with this model, and illustrate their approach on air quality (temporal) and precipitation (spatial) data.  

In another class of nested processes, conditional dependence is modeled via the covariance function. Consider, for example, the following nesting,
\begin{align*}
Y_n(\cdot) &= \GP(\xvec(\cdot)'\betab, C_n(\cdot,\cdot\,; \Yvec_{n-1}(\cdot))), \\
\Yvec_{n-1}(\cdot) &= \GP({\bf 0}, \Cmat_{n-1}(\cdot,\cdot\,; \Yvec_{n-2}(\cdot))), \\
\vdots ~~~~&= \quad \vdots \\
\Yvec_{1}(\cdot) &= \GP({\bf 0}, \Cmat_{1}(\cdot,\cdot)),
\end{align*}
where $Y(\cdot) \equiv Y_n(\cdot)$ is the underlying (often latent) process of interest and $\Yvec_1(\cdot),\dots,\Yvec_{n-1}(\cdot),$ are nested processes parameterizing the covariance functions of subsequent layers. \citet{Monterrubio_2020} consider the $n = 2$ case, where $Y_1(\cdot)$ is the (univariate) process describing the log of the length scale for the covariance function of $Y_2(\cdot)$, that is, $C_2(\cdot, \cdot\,; Y_1(\cdot))$, which is given by the non-stationary Mat{\'e}rn representation of \citet{Paciorek_2006}. \citet{Zhao_2021} call this model the ``batch deep Gaussian process regression model''. They also employ the \citet{Paciorek_2006} representation, but let $\Yvec_1(\cdot)$ be a bivariate Gaussian process, with the first variate the square root of the length scale, and the second variate the square root of the variance parameter. \citet{Zhao_2021} also present a dynamic state-space representation for the deep Mat{\'e}rn regression model for the \emph{temporal} case, based on known equivalencies between the two representations \citep[e.g.,][Chapter 12]{Sarkka_2019}. This representation is attractive as it allows sequential estimation methods (e.g., Kalman filtering based methods) to be used with deep hierarchies. This approach has yet to be applied in a spatio-temporal setting.

 \entry{State-space model}{A two-layer temporally-indexed model, where the first layer models the evolution of a latent state in time, and the second layer models the observations of the latent states\vspace{0.1in}}

 \entry{Kalman filter}{a sequential estimation method for linear, Gaussian, state-space models\vspace{0.1in}}

A related nested spatial process model is given by \citet{Chen_2021} in what they call ``DeepKriging''. Here, the bottom layer (referred to as an ``embedding layer'') of the model is given by the conventional multivariate spatial random effects model \citep[e.g.,][]{Nguyen_2017},
 
 \entry{Embedding layer}{a layer in a neural network which projects the input into a lower-dimensional space}

\begin{equation}\label{eq:DeepKriging1}
  \Yvec_1(\cdot) = \Wmat_{1,x}\xvec(\cdot) + \Wmat_{1,\phi}\phib(\cdot) + \bvec_1,
\end{equation}
where $\Wmat_{1,x}$ and $\Wmat_{1,\phi}$ are weight matrices that need to be estimated, $\xvec(\cdot)$ are covariates, $\phib(\cdot)$ are spatial basis functions, and $\bvec_1$ are bias parameters. The multivariate spatial process is then treated, up to a monotonic nonlinear transformation, as a set of basis functions for the subsequent layer. This approach yields the following nesting for $l = 2,\dots,n-1,$
$$
\Yvec_l(\cdot) = \Wmat_{l}\psi_{l-1}(\Yvec_{l-1}(\cdot)) + \bvec_l,
$$
where $\psi_{l}(\cdot)$ is the monotonic nonlinear transformation for the $l$th layer (applied element-wise) and  where the other quantities are defined similarly as in Equation \ref{eq:DeepKriging1}. The final layer (i.e., the process of interest) is then modeled as
$Y_n(\cdot) = \psi_n(\Wmat_{n}\psi_{n-1}(\Yvec_{n-1}(\cdot)) +  b_n).$ \citet{Chen_2021} show that the DeepKriging predictor is highly adaptable to non-stationary data, and that it can be quick to implement using GPU acceleration (see Section~\ref{sec:tech}). The DeepKriging architecture can be viewed as a special case of the deep generalized linear (mixed) model proposed by \citet{Tran_2020}.

\subsubsection{Deep hybrid spatio-temporal process models}\label{sec:spatiotemporal}

In this section we describe a few hybrid approaches for modeling the process component of dynamic spatio-temporal statistical models.  First, as noted in Section \ref{sec:tradAImethods}, RNNs have provided an effective way to model complex temporal dependency.  They have been used in the statistics context for time series applications \citep[e.g., see][for a hybrid RNN/stochastic volatility model]{Nguyen_2019}.   RNNs have also been combined in various ways with other neural architectures to accommodate spatial input \citep[e.g.,][]{dixon2019deep}.  As with multi-level BHM implementations of spatio-temporal processes, these implementations have a very large number of parameters and thus require a high volume of data and computational overhead to implement.  An alternative implementation of an RNN with a significantly more parsimonious representation is the echo state network \citep[ESN,][]{jaeger2001echo,jaeger2007echo}.  The ESN is a type of ``reservoir computing'' in which the hidden states and inputs evolve in a dynamical reservoir where the parameters (weights) that describe their evolution are drawn at random, with most assumed to be zero. Only parameters (weights) that are estimated at the output stage, that is, those that connect the hidden states to the output response, are estimated.

\citet{mcdermott2017ensemble} use this idea in a hybrid statistical/ESN model for spatio-temporal prediction with an additional quadratic output state. Their model is as follows: For time $t=1,\ldots,T$,
\begin{eqnarray}
\mbox{Response: }  & \; & \Zvec_t = \Vmat_1 \hvec_t + \Vmat_2 \hvec^2_t + \epsilonb_t, \quad \textrm{for} \quad \epsilonb_t \; \sim \; \textrm{Gau}(\zerob,\sigma^2_\epsilon \Imat), \label{eq:QESNresp} \\
\mbox{Hidden states: }  & \; &  \hvec_t = g_h\left(\frac{\nu}{|\lambda_w|}\Wmat \hvec_{t-1} + \Umat{\xvec}_t\right), \label{eq:QESNhidden}  \\
\mbox{Parameters: } & \; & \Wmat = [w_{i,\ell}]_{i,\ell}: w_{i,\ell} = \gamma^w_{i,\ell} \cdot \Unif(-a_w,a_w) + (1 - \gamma^w_{i,\ell}) \; \delta_0, \label{eq:QESNw} \nonumber \\
& \; & \Umat = [u_{i,j}]_{i,j}: u_{i,j} = \gamma^u_{i,j} \cdot \Unif(-a_u,a_u) + (1 - \gamma^u_{i,j}) \delta_0, \label{eq:QESNu}  \nonumber \\
& \; & \gamma_{i,\ell}^w \; \sim \; \Bern(\pi_w), \;\; \gamma_{i,j}^u \; \sim \; \Bern(\pi_u),  \nonumber 
\end{eqnarray}
where $\Zvec_t$ is the response vector at time $t$; $\hvec_t$ is the hidden state vector; $\xvec_t$ is a vector of input covariates; $\Wmat$ is the hidden-process-evolution weight matrix; $g_h(\cdot)$ is an activation function; $\Umat$ is the input weight matrix; and $\Vmat_1$, $\Vmat_2$ are weight matrices associated with the  linear and quadratic output, respectively.  In addition, $\delta_0$ is the Kronecker delta function at zero, $\lambda_w$ corresponds to the largest eigenvalue of $\Wmat$, and $\nu$ is an ESN control parameter.  The only parameters estimated in this formulation are $\Vmat_1$, $\Vmat_2$, and $\sigma^2_\epsilon$ from Equation \ref{eq:QESNresp}, obtained using a regularized regression (e.g., ridge or lasso).    Importantly, the elements of the matrices $\Wmat$ and $\Umat$ are drawn randomly as either a zero or uniformly in the range $(-a_w,a_w)$ and $(-a_u,a_u)$, respectively.  

 \entry{Reservoir Computing}{a type of machine learning where all hidden layer weights are chosen randomly and only the weights (parameters) associated with the output layer are learned}

\citet{mcdermott2017ensemble} used an ensemble approach to do uncertainty quantification with this model, by generating many copies of the hidden units $\hvec_t$ through different simulated weight matrices, and using the ensemble output to generate a predictive distribution.  \cite{bonas2021calibration} showed that one can ensure proper coverage in the ensemble approach through calibration. Although \citet{McDermott_2019a} demonstrated that one could use Bayesian inference to estimate the reservoir weight matrices in Equation \ref{eq:QESNhidden} with some computational effort, it only performed slightly better than the ensemble approach, and is therefore not worth the extra computational effort.  

It is straightforward to implement a spatio-temporal ESN in a deep context -- one simply allows the hidden states at one level to be the inputs to the next level.  Indeed, it has been demonstrated that such models are beneficial in that they can more readily utilize multi-scale temporal and spatial dependence in their predictions \citep{jaeger2007discovering, McDermott_2019b}.  Typically, as with CNNs, one performs some type of dimension reduction between layers to reduce the dimensionality of the hidden states and, ultimately, the number of variables used for prediction in the final layer.  For example, \citet{McDermott_2019b} give the following approach for time $t$: starting with the $n$th hidden layer that takes $\xvec_t$ as input, the model iterates from $l = n-1, \ldots, 1$ (note, the label ordering is opposite to that presented in Section \ref{sec:nested}, where here $n$ corresponds to the input layer): 
\begin{eqnarray}
   \text{Input Stage:} & \; &    \hvec_{t,n}=g_h \left(  \frac{\nu_n}{ |\lambda_{W_n}|} \Wmat_n \hvec_{t-1,n} + \Umat_n \xvec_t  \right) \label{eq:D-EESNinput}, \\
    \text{ Reduction Stage $l+1$:}   & \; & \widetilde{\hvec}_{t,l+1} \equiv \mathcal{Q}(\hvec_{t,l+1}), \;\;  l = n-1,\dots, 1, \label{eq:DEESNredStage} \\
  \text{Hidden Stage $l$:}   & \; &   \hvec_{t,l}=g_h\left(  \frac{\nu_l}{ |\lambda_{W_l}|} \Wmat_l \hvec_{t-1,l} + \Umat_l \widetilde{\hvec}_{t,l+1}  \right),  \;\;  l =  n-1,\dots,1 \label{hiddenStageOne} ,
\end{eqnarray}
where the weight matrices are generated randomly as above, and $\lambda_{W_l}, l = n,\ldots,1$ are the largest eigenvalues of their respective weight matrices,  $\nu_l$ are ESN control parameters (that are pre-specified), and $\mathcal{Q}(\cdot)$ is a dimension reduction function that reduces the dimension of $\hvec_{t,l}$ to $\widetilde{\hvec}_{t,l}$ for each level $l = n,\ldots,2$ (level 1 is typically not reduced, but could be). McDermott and Wikle (2019) use each of the dimension-reduced hidden states and the level 1 hidden states as possible predictors in the response level of the model (analogous to Equation \ref{eq:QESNresp}).  Note that the dimension reduction in Equation \ref{eq:DEESNredStage} can be unsupervised (e.g., principal component reduction or random projection) or supervised (e.g.,  with the use of autoencoders).  In the case of unsupervised dimension reduction, the hidden-state construction represented in Equations \ref{eq:DEESNredStage} and \ref{hiddenStageOne} is simply a multi-resolution stochastic transformation of the input.

As with the shallow spatio-temporal ESN presented above, uncertainty quantification can be accounted for by bootstrap ensemble approaches or through Bayesian inference. For example, the Bayesian approach given in \citet{McDermott_2019b} extends the basic output function and data stage of the ensemble deep spatio-temporal ESN.  Specifically, the ensemble member hidden states sampled from different random reservoirs are used in a regularized linear regression to model a transformation of the mean response from the data stage, similar to generalized additive models.  Specifically: 
\begin{align}
 \text{Data Stage:}  &  \qquad \Zvec_t | \alphab_t  \; \sim \; \textrm{Dist} ( \tilde{g}( \alphab_t), \Thetab ),  \label{eq:BD-EESNdata}  \\
 \text{Output Stage:}    & \qquad  \alphab_t =\frac{1}{n_{\res}} \sum\limits_{j=1}^{n_{\res}} \left[  \betab_1^{(j)}\hvec_{t,1}^{(j)} +  \sum\limits_{l=2}^n \betab_l^{(j)}  \widetilde{\hvec}_{t,l}^{(j)} \right] +\etab_t,  \label{eq:BD-EESNoutput}
 \end{align}
 where $\etab_t  \sim \text{Gau}(\zerob,\sigma^2_ \eta \Imat)$, ``Dist'' denotes an unspecified distribution (e.g., exponential family),  $\tilde{g}(\cdot)$ is some specified transformation (e.g., inverse link function), $\betab_{l}^{(j)}$ are regression matrices for $l=1,\ldots,n$ and the $j$th reservoir replicate, where $j=1,\ldots,n_{\res}$.  In the Bayesian implementation, stochastic search variable selection (SSVS) or other Bayesian variable selection methods can be used to regularize the regression matrices. 
 
The hybrid ESN approach, and variations thereof, have been used to successfully predict sea surface temperatures \citep{mcdermott2017ensemble}, soil moisture \citep{McDermott_2019b}, wind power \citep{Huang_2021},   industrial processes \citep{dixon2021industrial}, electricity prices \citep{klein2020deep}, asset volatility \citep{parker2021general}, and air pollution \citep{bonas2021calibration}.

\subsection{Deep learning for characterizing complex data models}\label{sec:data_model}

  \entry{Exponential transform}{the one-parameter transform given by $ g(Y(\cdot)) =(\exp(\lambda Y(\cdot)) - 1)/\lambda$ for $\lambda \ne 0$ and $g(Y(\cdot)) = Y(\cdot)$ for $\lambda = 0$\vspace{0.1in}}

It is common in various branches of statistics to develop probability models for a transformation of the data, rather than for the data themselves. Such transformations include the log transform, the exponential and power transforms \citep[e.g.,][]{Cressie_1978}, the Box--Cox transform \citep{Box_1964}, and the Tukey $g$-and-$h$ transform \citep{Tukey_1977}. Many of these have been used in spatial statistics \citep[e.g.,][]{deOliveira_1997, Xu_2017}. These transformations generally have a parsimonious parameterization, and are relatively simple in form. More flexibility can be achieved by expressing the transformation using a (suitably structured) deep learning architecture.

The ``warped GP'' of \citet{Snelson_2004} is given by   
\begin{align}
  g_\varthetab(Z_i) &= Y(\svec_i) + \epsilon_i,\quad i = 1,\dots,m, \\
  Y(\cdot) &\sim \GP(\mu(\cdot), C(\cdot,\cdot)),
\end{align}
\entry{Box--Cox transform}{the one-parameter transform given by $ g(Y(\cdot)) =(Y(\cdot)^\lambda - 1)/\lambda$ for $\lambda \ne 0$ and $g(Y(\cdot)) = \log Y(\cdot)$ for $\lambda = 0$}
where $\{Z_i\}$ are the observations, $\{\epsilon_i\}$ are the measurement errors, $g_\varthetab(\cdot)$ is a monotonic function parameterized via $\varthetab$ and, in the context of spatial statistics, $Y(\cdot)$ is  the spatial process of interest with mean function $\mu(\cdot)$ and covariance function $C(\cdot,\cdot)$. \citet{Snelson_2004} propose using a one-layer net of $\tanh(\cdot)$ functions to model $g_\thetab(\cdot)$. \citet{Rios_2019} extend the warped GP to the ``compositionally warped GP'' by expressing $g_\varthetab(\cdot)$ as a (deep) composition of elementary functions which have explicit derivatives and inverses; these include the Box--Cox transform and the Sinh-Arcsinh transform. \citet{Murakami_2021} use the compositionally warped GP in a spatial mixed-model setting, while \citet{Maronas_2021} propose a computationally-efficient variational algorithm for fitting the model. Note that, unlike in conventional generalized linear models (GLMs), no distribution is pre-specified for the $\{Z_i\}$, which depends on $\varthetab$ that needs to be estimated. This approach is thus distinct from the deep GLMM setting of \citet{Tran_2020} where the $\{Z_i\}$ are assumed to come from a known exponential family and the link function is fixed, and where a deep net is used to model what is conventionally the linear component. The latter, however, bears connections to the spatial GLMM of \citet{Diggle_1998} and thus is a strong candidate for use in spatial applications.  Recently, \citet{bradley2022joint} considered a multi-level BHM for unknown transformations of multiple response-type data.  Their approach takes into account the uncertainty associated with the unknown transformation and has been applied to both spatial and spatio-temporal data.

  \entry{Sinh-Arcsinh transform}{the two-parameter transform given by $g(Y(\cdot)) = \sinh(\lambda\sinh^{-1}Y(\cdot) - \gamma)$}

\subsection{Deep learning for parameter estimation in spatial and spatio-temporal empirical hierarchical models}

As discussed in Sections \ref{sec:tradspace} and \ref{sec:tradST}, classical hierarchical spatial and spatio-temporal statistical models generally involve parameters that characterize the first-order, second-order, and sometimes third and higher order, properties of the process. Estimating these parameters is often a computational bottleneck for a variety of reasons. Sometimes the likelihood is difficult to evaluate, especially as the size of the data set increases. In other cases the likelihood may be computationally tractable, but difficult to explore using conventional optimization techniques. This problem has led to a recent drive in the spatial statistics literature to employ neural networks to construct mappings between the observation space and the parameter space. Once trained, such a network is, in principle, able to provide practically-usable parameter estimates from any observed data set in a fraction of the time needed by conventional (e.g., likelihood) techniques, irrespective of the model complexity.

Recall the IDE dynamic spatio-temporal model discussed in Section~\ref{sec:spatiotemporal}.  One of the greatest challenges with IDE models is estimating the parameters associated with the mixing kernel, especially when they are parameterized in a highly flexible manner (e.g., to allow for spatio-temporally-varying mixing). In particular, joint inference over the latent spatio-temporal process and a spatio-temporally-varying mixing kernel is notoriously difficult and time consuming. \citet{Zammit_2020} proposed to alleviate the computational burden by finding a (highly complex) time-invariant map between the mixing kernel parameters and lagged values of the process.  This complex map was described using a CNN as in \citet{deBezenac_2019}, and was fitted offline using a copious amount of reanalysis geophysical data. \citet{Zammit_2020} then converted Equation  \ref{eq:DSTM_IDE} into a state-dependent (and hence nonlinear) IDE, for which standard Kalman-based filtering techniques could be used. Their results showed a 100-fold decrease in the time required to make inference on the process and the parameters over standard moving-window-based maximum-likelihood methods. They also demonstrated that their approach has robust transfer learning potential by generating successful forecasts of a process (precipitation) that is considerably different from that on which the CNN was trained (sea surface temperature).

\citet{Gerber_2021} used a similar CNN architecture to that of \cite{Zammit_2020} to estimate the length scale and the effective degrees of freedom (in their case a variance parameter) of a Gaussian process with a Mat{\'e}rn covariance function observed in noise. The CNN was trained using thousands of simulated fields, corresponding to different parameters, as input data. The parameters the fields were simulated at were then used as output data. \citet{Gerber_2021} showed that their CNN estimator was comparable to the maximum likelihood estimator in terms of bias and variance and, like \citet{Zammit_2020}, reported a hundred-fold speed up in estimation. \citet{Lenzi_2021} considered a similar CNN framework for estimating parameters in models of spatial extremes. This approach to parameter estimation is still in its infancy in spatial statistics, but has seen wide use in a variety of related areas that require parameter estimation. For example, \citet{Rudi_2021} use a similar approach for estimating the parameters of a system of ordinary differential equations.

Deep networks have also been used to facilitate inference of spatial models with intractable likelihoods. \citet{Vu_2022}, for example, used deep compositional spatial models to emulate the sufficient statistics required to construct synthetic likelihood functions. These synthetic likelihood functions were then used to speed up parameter inference with the spatial Potts model and the spatial autologistic model.

  \entry{Potts and autologistic models}{Spatial models on a lattice where each vertex may be in one of several (finite) states}

\section{Other uses of deep learning in spatial and spatio-temporal statistics}\label{sec:other_uses}

Section~\ref{sec:DeepStat} showcased instances of classical spatial and spatio-temporal models that incorporate ideas or model formalisms that are often seen in the deep learning literature. In this section, we focus on specific types or applications of spatial and spatio-temporal models (specifically, point-process models, emulation, and reinforcement learning) that at their core incorporate deep learning architectures.

\subsection{Deep Poisson point process models}

\subsubsection{Measure transport for modeling non-homogeneous Poisson point processes}\label{sec:pp}

\citet{Tabak_2010} introduced the concept of a \emph{flow} for constructing complicated probability density functions, which has seen substantial use and development in the machine learning literature \citep[e.g.,][]{Rezende_2015}. Consider some continuous and differentiable density function $f_0(\xvec), \xvec \in \Xset,$ which one wishes to model, and let $T_\thetab(\cdot)$ be some bijective and differentiable map parameterized through the parameter vector $\thetab$. Let $f_1(\cdot)$ be a \emph{reference} density that is easy to evaluate. In this setting, one expresses $f_0(\cdot)$ via the popular change of variables formula
\begin{equation}\label{eq:changeofvars}
f_0(\xvec) =  f_1(T_\thetab(\xvec))|\det(\nabla(T_\thetab(\xvec)))|, \quad \xvec \in \Xset.
\end{equation}
The attraction of this construction is that one need only estimate $\thetab$ in order to construct $f_0(\cdot)$, which can be done relatively efficiently through likelihood methods. Specifically, assume that one has a sample $\xvec_i, i = 1,\dots,N$; then one could obtain an estimate $\hat\thetab$ for $\thetab$ via the operation,
$$
\hat\thetab = \argmax_\thetab\left\{\sum_{i=1}^N \log f_1(T_\thetab(\xvec_i))  + \log \left| \det(\nabla(T_\thetab(\xvec_i))) \right|   \right\}.
$$
Since bijection and differentiability are preserved through composition, it is common to string together several transformations through composition, $K$ say, to obtain a more complex mapping (this stringing together is what gives rise to the term ``flow''), so that $T_{\thetab}(\cdot) \equiv T_{K,\thetab} \circ \cdots \circ T_{1,\thetab}(\cdot)$, where each of $T_{1,\thetab}(\cdot),\dots T_{K,\thetab}(\cdot)$ is bijective and differentiable. Various models for $T_{k,\thetab}(\cdot), k = 1,\dots,K,$ have been proposed that, while being bijective and differentiable, also lead to a Jacobian determinant in Equation \ref{eq:changeofvars} that is computationally tractable. Pertinent to this review are those constructed via triangular maps, where the $i$th output of $T_{k,\thetab}(\xvec)$, that is, $T^{(i)}_{k,\thetab}(\xvec)$, is a monotonic, nonlinear, function of the $i$th dimension of $\xvec$ (i.e., $\xvec^{(i)}$), with parameters that often depend in a highly nonlinear manner on $\xvec^{(1)} ,\dots,\xvec^{(i-1)}$ via a deep learning framework \citep[e.g.,][]{Kingma_2016, Papamakarios_2017, Huang_2018}.

Normalizing flows, as they are often called since $f_1(\cdot)$ is typically the normal probability density function, are of interest in spatio-temporal statistics for various reasons. First, they could be used to improve variational inference when dealing with large spatial data \citep[e.g.,][]{Hensman_2013, Rezende_2015}. More directly though, one can use normalizing flows to model the intensity function of a non-homogeneous (temporal, spatial, or spatio-temporal) Poisson point process. The connection arises from the fact that if  $\lambda(\cdot)$ is the intensity function of a Poisson point process on $\Xset$, and if $\mu_\lambda(\Xset) \equiv \int_\Xset\lambda(\xvec)\intd\xvec$ is the integrated intensity, then $\lambda(\cdot)/\mu_\lambda(\Xset)$ is a density, referred to as the process density by \citet{Taddy_2010}. \citet{Ng_2022A} propose using the neural autoregressive flow of \citet{Huang_2018} for modeling the process density, and estimate the integrated intensity as the number of observed points, $N$. They show that, under mild regularity assumptions on the intensity function, the neural autoregressive flow is a universal approximator (i.e., that it can model arbitrarily complex intensity functions). \cite{Ng_2022B} apply a similar approach to point processes on the sphere using exponential map radial flows \citep{Sei_2013, Rezende_2020}.
 
  \entry{Non-homogeneous Poisson point process with intensity function $\lambda(\cdot)$}{a random finite subset of a domain $G$, where the number of points in any $A \subset G$ is Poisson distributed with parameter $\int_A \lambda(\svec)\intd\svec$, and where the number of points in disjoint regions are independent}

\subsubsection{Conditional intensity function modeling}

A considerable literature has recently emerged on the modeling of conditional intensity functions using ordinary differential equations that are built using neural networks. These intensity functions, which are conditional on the event history, are often more applicable to real world processes, such as financial or crime data. \citet{Jia_2019} apply the neural ordinary differential equations of \citet{Chen_2018} to model conditional intensity functions, while \citet{Chen_2020} extend the concept to the spatio-temporal case. 

\citet{Zhu_2020} take a different approach and directly model the influence of a past point on the intensity function using a Gaussian mixture, where the parameters in each mixture component are the outputs of a simple, one-layer, neural net that take the spatial coordinates as input. The model yields a flexible conditional-intensity function that is highly spatially heterogeneous and applicable in various contexts such as the analysis of earthquake data and crime data.

\subsection{Deep emulation}

An emulator is a model that acts as a surrogate for a more complex, generally numerical and physics-based, model. Emulators are most often forward maps between the (input) parameter space and the (output) response space of the numerical model, although emulators for inverse maps have also been developed (see below). They tend to be of greatest use when the numerical model is computationally intensive to run, and when the output varies reasonably smoothly with small changes in the input. Emulators find most use in prediction (i.e., for predicting numerical model output for as yet unprobed parameters), experimental design, and calibration (i.e., for tuning numerical models to observed data).

The most common emulator is the Gaussian process emulator \citep{Kennedy_2001}, where the map between the  input parameters and the numerical model outputs is modeled via a Gaussian process. Several developments on the vanilla Gaussian process have been proposed for the purpose of emulation, such as the treed Gaussian process \citep{Gramacy_2008} and the deep Gaussian process  of \citet{Damianou_2013} or variants thereof \citep{Monterrubio_2020, Ming_2021, Marmin_2022, Sauer_2022}.

  \entry{Treed Gaussian process}{a nonstationary Gaussian process model constructed by modeling several stationary Gaussian process models on a random, marginalized, partitioning of the spatial domain}

The output of a numerical, physics-based, model is often temporal, spatial, or spatio-temporal, and several approaches have been developed to deal with this extra level of complexity. \cite{Leeds_2013} used random forests to model the three-dimensional output of a biogeochemical model,  while \citet{leeds2014emulator} considered a deep quadratic nonlinear model to emulate a multivariate spatio-temporal process.  \citet{Zhang_2015} used a nonseparable Gaussian process on the parameter-spatial-temporal input space   to model the output of complex computational fluid dynamic models; see also \citet{Castruccio_2014} and \citet{Chang_2016} for related examples.

  \entry{Random forest}{A collection of decision or regression trees, trained on random subsets of the data and input variables}

Research into the use of deep learning models for emulating the spatial or spatio-temporal output of numerical models is in its infancy, but has a lot of potential benefit that will likely make it an active area of research in the coming years. This benefit is seen in the work of \citet{Bhatnagar_2022} who use deep learning models known as long-short term memory (LSTM) models to find a complex map between the numerical model output and the input parameters used to generate the output (i.e., an inverse map); this model can then be directly used for calibration. Another example is the work of \citet{Cartwright_2021}, who emulate the spatial output of a Lagrangian particle dispersion model (LPDM, which simulates particle trajectories in the atmosphere from a source location) using a deep learning model known as a convolutional variational autoencoder (CVAE). The authors show that the CVAE can be used to effectively predict the output of the LPDM over a wide spatial domain given only a few simulations, and that the CVAE considerably outperforms the conventional emulator based on singular vectors \citep[e.g.,][]{Hooten_2011}. Recently, \citet{gopalan2021higher} extended the singular vector approach to higher-order tensor decompositions to emulate complex multi-dimensional spatio-temporal data, including the movement trajectories of agents in an agent-based model. Their approach is flexible in that different machine learning methods (e.g., random forests and neural networks) or GP regression models can be used in the various tensor dimensions.

  \entry{Convolutional variational autoencoder}{An autoencoder where the encodings are defined as random variables, where training is done using variational Bayes, and where the encoder and decoder layers are convolutional layers}

\subsection{Reinforcement learning}

Reinforcement learning (RL) is goal-oriented learning from the interaction between agents and their environment, where the agents learn to take actions that maximize a specified reward function.  The RL framework expands the characterization of agents from traditional agent-based models to include notions of perception and memory, where perception is related to the agent’s state, and memory is incorporated by allowing control parameters to be learned based on the agents' experience with their environment \citep[see][for a classic overview]{sutton1998introduction}. RL has seen a resurgence due to the success of embedding deep models in various components of the learning process \citep[see the overview in][]{ henderson2018deep}. Given that many agent-based systems, such as autonomous vehicle control systems and collective animal movement, are formulated in space and time, it is natural to consider RL for such problems \citep[e.g.,][]{ma2021reinforcement, tampuu2017multiagent}.  However, for many such systems it is challenging to define the local costs or rewards that control agent behavior {\it a priori}, which has led to interest in inverse reinforcement learning (IRL), whereby one uses observed system behavior to learn the underlying costs or rewards \citep[e.g.,][]{ng2000algorithms}. In the spatio-temporal statistical context, \citet{schafer_2020} used Bayesian IRL to recover  the costs that guppies in a tank consider with respect to the tradeoffs that exist between collective movement and movement to a safe zone.

\entry{Agent-Based Modeling}{a bottom-up modeling approach that specifies simple rules for how agents interact with each other and their environment; these rules lead to complex behavior }

\section{Technologies used for deep learning in spatial statistics}\label{sec:tech}

Most of the methods and techniques discussed in Sections \ref{sec:tradAI}--\ref{sec:other_uses} require the estimation of parameters appearing in a deep hierarchy, generally via the optimization of a (regularized) likelihood function. The past decade has seen a dramatic increase in the availability of tools and hardware specifically designed to solve this problem. The two most important ones, which are indispensable for anyone implementing deep learning models in spatial statistics, are deep learning software libraries, and graphics processing units (GPUs).

At the time of writing, the two most popular deep learning libraries are \texttt{PyTorch} and \texttt{TensorFlow/Keras}. Both are open source Python libraries that greatly facilitate model construction and fitting through various features. First, they allow one to construct large deep learning models with ease; for example, creating a convolutional layer in a CNN would only require calling a single function. Second, they both offer functionality for automatic differentiation \citep[e.g.,][]{Paszke_2017}, so that derivatives during optimization can be obtained quickly and effortlessly. Third, they offer a wide-range of (stochastic) gradient-descent strategies (also referred to as (stochastic) optimizers) that have been proven useful for these models, such as Adagrad \citep{Duchi_2011} and Adam \citep{Kingma_2014}. Finally, they both offer seamless GPU integration, which are a practical requirement when training large, deep learning, models. Both \texttt{PyTorch} and \texttt{TensorFlow/Keras} functionality have been made available to \texttt{R} users \citep{R}; see, for example, \url{https://tensorflow.rstudio.com/}.

  \entry{Autodifferentiation}{a tool to automatically compute the derivatives of a computation by application of the chain rule on the elementary operations from which it is constructed}

Each likelihood evaluation when training a model incorporating a deep hierarchy generally requires a substantial amount of high-dimensional, but relatively simple, matrix calculations that are na{\"i}vely parallelizable (such as addition and multiplication). GPUs contain an exorbitant number of processing cores: In the year 2022 a high-end GPU contained several thousands of cores while a typical high-end central processing unit (CPU) contained a few dozen compute cores. GPUs are thus poised to take advantage of the parallelizable matrix operations, and offer a drastic computational benefit over conventional CPUs.  They are also typically available with a very large memory bandwidth (on the order of terabytes per second, as opposed to gigabytes per second) so that the large portions of data in memory that are required for computation can be quickly accessed. GPUs also offer considerable improvement in compute speed, even when shallow models are used; see \url{https://hpc.niasra.uow.edu.au/azm/Spatial_GPUs_TFv2.html} for a spatial modeling experiment where a GPU is used to fit a (shallow) spatial model nearly 50 times faster than a CPU, although both are using the same optimizer and carrying out the same calculations.

Section~\ref{sec:DeepStat} reviewed several statistical deep learning/hierarchical models that have been adopted for analyzing spatial data. Software for implementing these methods is still in its infancy, and largely in the form of `reproducible software' accompanying journal articles. Yet, such software is also an excellent starting point for exploring the features and implementation of these deep learning models, and are a valuable resource. At the time of writing, software for DeepKriging was available at \url{https://github.com/aleksada/DeepKriging}, software for fitting deep GMRFs was available at \url{https://bitbucket.org/psiden/deepgmrf/src/master/}, and software for fitting deep compositional spatial models was available at \url{https://github.com/andrewzm/deepspat}. 

\section{Conclusion}\label{sec:conclusion}

In this overview, we presented a statistician's perspective and contemporary snapshot of deep learning for spatial and spatio-temporal data.  We gave a brief overview of traditional statistical and deep learning models for such data, noting that  ``deep'' models have been integral to modeling spatial and spatio-temporal data in statistics since the 1990s, when computational approaches for fitting multi-level (deep) Bayesian hierarchical models became available.  Our focus in this review was on hybrid machine-learning/statistical models that utilize deep learning and that can still accommodate uncertainty quantification and some measure of explainability or interpretability.  In the context of latent processes, we discussed hybrid methods such as deep Gaussian processes and deep echo state networks.  We also discussed how deep models can be used to characterize complex data models in more traditional multi-level statistical models. We then presented some recent work that has used deep learning to estimate the parameters of various statistical models, from those appearing in covariance functions of spatial models, to those characterizing the transition operator in a spatio-temporal dynamical model.  We proceeded to present some other examples  where deep models were used in the context of spatial and spatio-temporal data, namely in point process modeling, computer model emulation, and reinforcement learning.  Finally, we concluded with a brief overview of technologies that have enabled deep learning at large, and that will be indispensable to the practitioner aspiring to do deep learning with spatial and spatio-temporal data.

Although the fusion of deep machine learning and statistical approaches for spatial and spatio-temporal data is in its infancy, there is substantial research interest in these methods.  In addition to extensions and implementations of the methods described here, there are several areas that will likely see greater exploration in the near future.  These include the increased use of novel stochastic optimization algorithms, the development of methods for covariance free spatial and spatio-temporal prediction, the development of new explainability and interpretability methods, the incorporation of multi-type, multi-support data, and the development of efficient ways to specify optimal deep architectures.  Deep learning itself is a growing area in machine learning/computer science, and as new methods are developed, they will almost certainly be used as inspiration for enhancing traditional statistical methods for analyzing spatial and spatio-temporal data.  

\section*{Acknowledgements}

Christopher K. Wikle's research was supported by the U.S.~National Science Foundation (NSF) grant SES-1853096.  Andrew Zammit-Mangion’s research was supported by an ARC Discovery Early Career Research Award, DE180100203. The authors would like to thank Yi Cao, Wanfang Chen, and Per Sid{\'e}n, for help with implementing and running software for DeepKriging and deep GMRFs.

\bibliographystyle{deep_spat_ARSIA} 
\bibliography{deep_spat_ARSIA}

\end{document}